\documentclass[runningheads]{llncs}
\usepackage[T1]{fontenc}
\usepackage{graphicx,verbatim}
\usepackage{color}
\usepackage{hyperref}

\usepackage{amsmath}
\usepackage{multirow}
\usepackage{amsfonts}
\usepackage{amssymb}
\usepackage{tabularx} 
\usepackage{soul}
\usepackage{marvosym}

\begin{document}

\title{Interpretable Probabilistic Medical Image Segmentation via Gaussian Process with Explicit Modelling of Annotation Bias and Variability}
\titlerunning{Interpretable Probabilistic Segmentation using SVGP}
%

\author{
Qi Li\inst{1}\textsuperscript{(\Letter)}\and
%
Yuliang Huang\inst{1} \and
Shaheer U. Saeed\inst{1,2} \and
Qianye Yang\inst{1,3} \and
Vasilis Stavrinides\inst{4,5,6} \and
Zachary M. C. Baum\inst{1} \and
Dean C. Barratt\inst{1} \and
J. Alison Noble\inst{3} \and
Tom Vercauteren\inst{7} \and
Yipeng Hu\inst{1}}

\authorrunning{Q. Li et al.}
\institute{UCL Hawkes Institute, Department of Medical Physics and Biomedical Engineering, University College London, London, U.K.\\
\email{qi.li.21@ucl.ac.uk}\and
Centre for Bioengineering, Digital Environment Research Institute, School of Engineering and Materials Science, Queen Mary University of London, London, U.K. \and
Institute of Biomedical Engineering, Department of Engineering Science, University of Oxford, Oxford, U.K. \and
UCL Cancer Institute, University College London, London, U.K.\and
Department of Urology, University College London Hospitals NHS Trust, London, U.K. \and
Department of Radiology, Imperial College Healthcare, London, U.K. \and
School of Biomedical Engineering \& Imaging Sciences, King’s College London, London, U.K.
}

\maketitle              
\begin{abstract}
Deep learning-based medical image segmentation models are trained using annotations that exhibit systematic bias and variability across raters. While probabilistic multi-rater approaches can emulate annotator-specific delineations, annotator characteristics are typically encoded implicitly in deep latent feature space, making direct analysis of their influence on predictive distributions less straightforward.
We propose a logit-space probabilistic segmentation framework based on stochastic variational Gaussian Process that explicitly decomposes predictions into an image-dependent reference logit distribution and annotator specific perturbations parameterised by bias and variance. This formulation enables more explicit analysis on how intra- and inter-rater variability propagate to predictive distributions.
We evaluate the method on a multi-annotator medical image dataset, which shows that explicitly modelling annotator specific perturbations improves uncertainty calibration while maintaining comparable segmentation accuracy, compared with state-of-the-art multi-rater probabilistic segmentation method. The learned bias and variance parameters quantitatively reflect annotator-specific behaviour. Furthermore, controlled perturbation experiments over bias and variance demonstrate how changes in annotator parameters systematically influence predictive performance.
The code used in this paper is made publicly available at \url{https://github.com/QiLi111/GPS-Var}.

\end{abstract}

\section{Introduction}
\label{Introduction}

Deep learning-based segmentation has become a cornerstone of medical image analysis, supporting applications ranging from diagnosis~\cite{antonelli2022medical} and treatment planning~\cite{zaidi2010pet} to image-guided interventions~\cite{grammatikopoulou2021cadis}. The success of these models critically depends on the availability of high quality annotated data~\cite{wang2021annotation}. However, in medical imaging, annotations are often subjective and heterogeneous even among experienced experts~\cite{ji2021learning}. As a result, the supervision signals inevitably contain systematic biases and variability that propagate into learned models~\cite{sylolypavan2023impact,campagner2021ground}, yet how these factors influence predictive outputs is not always explicitly characterised.

In response to annotation variability, prior work has developed probabilistic segmentation frameworks~\cite{kohl2018probabilistic,kohl2019hierarchical,baumgartner2019phiseg} and multi-rater learning approaches~\cite{hu2019supervised,schmidt2023probabilistic,zepf2023that}. These methods effectively capture inter-observer disagreement, typically by conditioning latent representations on annotator identity or by modelling annotations as noisy transformations of an underlying consensus~\cite{zhang2020disentangling,rodrigues2018deep}. While such approaches successfully emulate annotator-specific delineations, the influence of annotator-dependent noise on predictive probabilities is often mediated through deep nonlinear mappings, making analytical examination of how bias and variability propagate to output less direct.

Understanding this propagation can inform dataset curation strategies under fixed labelling budgets, where one must decide whether to prioritise consensus training, repeated annotations, or increasing number of raters, etc. Moreover, in safety-critical clinical settings, it is essential to distinguish uncertainty arising from intrinsic image ambiguity~\cite{kendall2017uncertainties} from that induced by imperfect supervision, as these sources require different mitigation strategies. Existing frameworks can represent these uncertainty sources implicitly, but explicit formulations that allow direct sensitivity analysis with respect to annotator bias and variability remain limited.

\begin{figure}
\includegraphics[width=\textwidth]{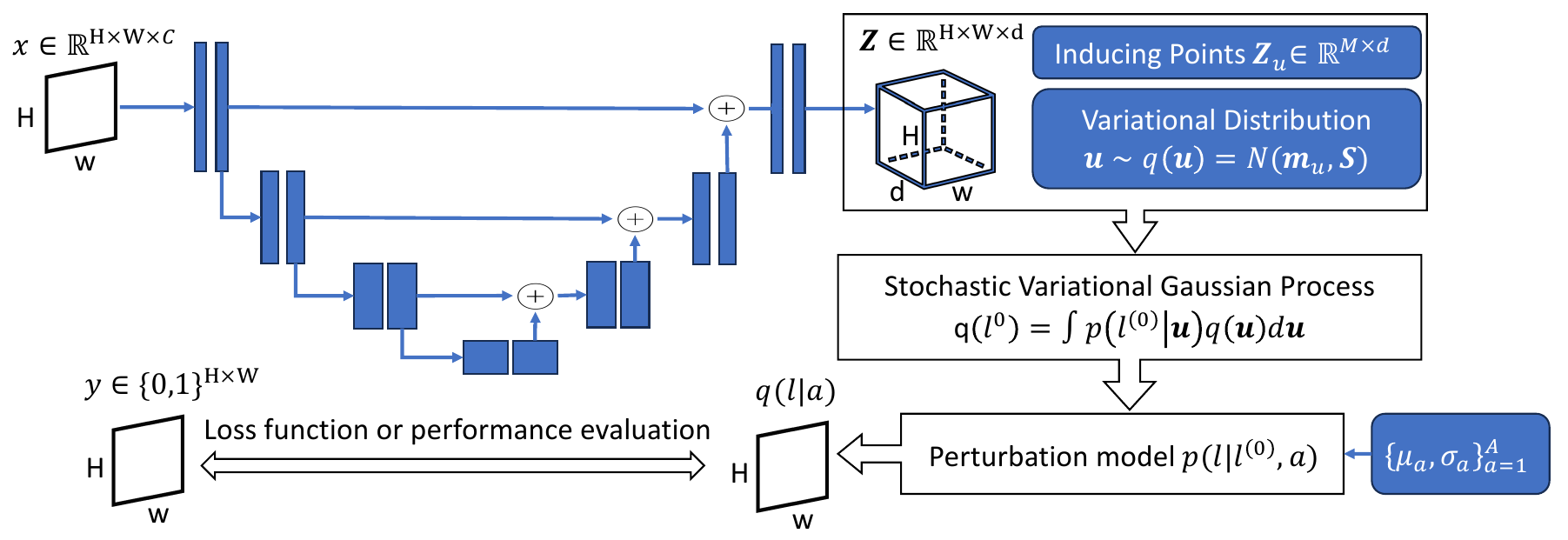}
\caption{An input image is encoded by a U-Net into latent features, on which an SVGP~\cite{hensman2015scalable} models the reference logit distribution. Annotator-specific bias and variability are then applied as perturbations. Blue blocks indicate trainable parameters. Definition of relevant notations is given in Section \ref{method}.}
\label{overview}
\end{figure}

In this work, we propose a logit-space reformulation of multi-annotator segmentation that makes annotator-specific bias and variability explicit in the predictive distribution. Image-dependent predictions are represented by a reference logit distribution that reflects the segmentation implied purely by image content under ideal supervision, while observed annotations are modelled as additive perturbations parameterised by annotator-specific bias and variance. 
We estimate the reference logits using a stochastic variational Gaussian Process (SVGP)~\cite{hensman2015scalable,pan2024uncertainty}. Under this formulation, a closed-form rater-conditioned predictive probability is guaranteed, allowing sensitivity analysis of segmentation performance with respect to annotator bias and variability.
An overview of the framework is shown in Fig.~\ref{overview}.

\section{Method}\label{method}

\subsection{Logit-Space Reformulation with Annotator Modelling}

Let $x \in \mathbb{R}^{H \times W \times C}$ denote an input image and $y=\{y_i\}_{i=1}^{P}$ its pixel-wise segmentation mask, where $i$ indexes spatial locations and $P=H\times W$. Let $a$ denote the annotator index.
We consider the predictive distribution conditioned on annotator $a$:
\begin{equation}
p(y \mid x, a)
=
\int p(y \mid l)\, p(l \mid x, a)\, dl,
\end{equation}
where $l=\{l_i\}_{i=1}^{P}$ are pixel-wise logits.
Conditioned on logits, labels are assumed independent across pixels:
$
p(y \mid l)
=
\prod_{i=1}^{P} p(y_i \mid l_i),
$
where $p(y_i \mid l_i)$ denotes a classification likelihood (e.g. Bernoulli likelihood with sigmoid function).

To separate intrinsic image variability from annotator effects, we introduce reference logits $l^{(0)}=\{l_i^{(0)}\}_{i=1}^{P}$, representing the logits under ideal (noise-free and unbiased) supervision. Using this intermediate representation, we factorise
\begin{equation}
p(l \mid x, a)
=
\int
p(l \mid l^{(0)}, a)\,
p(l^{(0)} \mid x)\,
dl^{(0)}.
\end{equation}
We model annotator-induced perturbations as
\begin{equation}
p(l \mid l^{(0)}, a)
=
\prod_{i=1}^{P}
\mathcal{N}\!\big(
l_i^{(0)} + \mu_a,\;
\sigma_a^2
\big),
\end{equation}
where $\mu_a \in \mathbb{R}$ captures systematic bias and $\sigma_a^2 \in \mathbb{R}^{+}$ models variability of annotator $a$. Under this formulation, $p(l^{(0)} \mid x)$ captures intrinsic uncertainty caused by image ambiguity, while $(\mu_a,\sigma_a^2)$ explicitly quantify supervision-induced shift and dispersion, enabling analysis of how annotator bias and variability propagate through the logit distribution and affect predictions.

\subsection{Deep Kernel Gaussian Process for Reference Logits}

Assume the training set contains $V$ images, each with $P$ pixels, yielding a total of $N = VP$ training inputs for the Gaussian Process~\cite{williams2006gaussian}.

Let $\phi_\psi : \mathbb{R}^{H \times W \times C}
\rightarrow
\mathbb{R}^{H \times W \times d}$ denote a convolutional feature extractor parameterised by $\psi$. Given image $x_v$, the resulting feature map is $\phi_\psi(x_v)$. For pixel $i$ in image $v$, we define the feature vector $\mathbf{z}_{v,i} = \phi_\psi(x_v)_i \in \mathbb{R}^{d}$, and collect all training features as $\mathbf{Z} = \{\mathbf{z}_{v,i}\}_{v=1,i=1}^{V,P}
\in \mathbb{R}^{N \times d}$.

We model the reference logit as $l^{(0)}_{v,i} = f(\mathbf{z}_{v,i})$, where the latent function $f$ follows a zero-mean Gaussian Process prior $f(\cdot) \sim \mathcal{GP}(0, k_{\varphi}(\cdot,\cdot))$. Here, $k_{\varphi} : \mathbb{R}^{d} \times \mathbb{R}^{d} \to \mathbb{R}$ is a symmetric positive semi-definite kernel applied on the latent features, parameterised by hyperparameters $\varphi$ \cite{wang2021learning,wilson2016deep}. For example, an RBF kernel \cite{bishop2006pattern} takes the form $k_{\varphi}(\mathbf{z}_i,\mathbf{z}_j)
=
\sigma_f^2
\exp(-\|\mathbf{z}_i-\mathbf{z}_j\|^2/(2\ell^2))$, where $\varphi = \{\sigma_f, \ell\}$.

Stacking all function values as $\mathbf{f}=\{f(\mathbf{z}_{v,i})\}_{v,i} \in \mathbb{R}^{N}$, the GP prior implies
\begin{equation}
\mathbf{f}
\sim
\mathcal{N}(\mathbf{0}, \mathbf{K}_{NN}),
\end{equation}
where $(\mathbf{K}_{NN})_{ij}
=
k_{\varphi}(\mathbf{z}_i,\mathbf{z}_j)$.

\subsection{Stochastic Variational Gaussian Process}

Since $N = VP$ can be large, exact GP inference scales cubically in $N$ \cite{snelson2005sparse,titsias2009variational} and becomes computationally infeasible. We therefore adopt the stochastic variational Gaussian Process (SVGP) \cite{hensman2013gaussian,hensman2015scalable} with $M \ll N$ inducing inputs.
Let $\mathbf{Z}_u = \{\mathbf{z}_u^{(j)}\}_{j=1}^{M}
\in \mathbb{R}^{M \times d}$ denote inducing features and $\mathbf{u}=f(\mathbf{Z}_u)\in\mathbb{R}^{M}$ the corresponding inducing variables. Under the GP prior, $(\mathbf{f},\mathbf{u})$ are jointly Gaussian.
SVGP further introduces a variational distribution $q(\mathbf{u})=\mathcal{N}(\mathbf{m}_u,\mathbf{S})$ to approximate the intractable posterior $p(\mathbf{u} \mid \mathbf{y})$. For brevity, we make the conditioning on input images $x$ implicit in the notation.

For an input feature $\mathbf{z}_*$, the predictive distribution over the corresponding reference logit $l^{(0)}_*=f(\mathbf{z}_*)$ satisfies
\begin{equation}
p(l^{(0)}_* \mid \mathbf{y})
=
\int
p(l^{(0)}_* \mid \mathbf{u})\,
p(\mathbf{u} \mid \mathbf{y})
\, d\mathbf{u}.
\end{equation}
The conditional $p(l^{(0)}_* \mid \mathbf{u})$ can be derived from the joint Gaussian distribution of $l^{(0)}_*$ and $\mathbf{u}$ under the GP prior.
Replacing the true posterior with its variational approximation $p(\mathbf{u} \mid \mathbf{y}) \approx q(\mathbf{u})$ yields
\begin{equation}
p(l^{(0)}_* \mid \mathbf{y})
\approx
\int
p(l^{(0)}_* \mid \mathbf{u})\,
q(\mathbf{u})
\, d\mathbf{u}
=
\mathcal{N}(\tau,\eta^2),
\end{equation}
where the predictive mean and variance are
\begin{align}
\tau
&=
\mathbf{K}_{*M}
\mathbf{K}_{MM}^{-1}
\mathbf{m}_u, \\
\eta^2
&=
K_{**}
+
\mathbf{K}_{*M}
\mathbf{K}_{MM}^{-1}
(\mathbf{S}-\mathbf{K}_{MM})
\mathbf{K}_{MM}^{-1}
\mathbf{K}_{M*}.
\end{align}
Here,
$\mathbf{K}_{MM}=k_{\varphi}(\mathbf{Z}_u,\mathbf{Z}_u)$,
$\mathbf{K}_{*M}=k_{\varphi}(\mathbf{z}_*,\mathbf{Z}_u)$,
and $K_{**}=k_{\varphi}(\mathbf{z}_*,\mathbf{z}_*)$.

\subsection{Predictive Distribution and Model Training}

Given $A$ annotators, let $l_{*,a}$ denote the predicted logit for a pixel conditioned on annotator $a$. Under the proposed model,
\[
l_{*,a} \mid l_*^{(0)}
\sim
\mathcal{N}(l_*^{(0)}+\mu_a,\sigma_a^2),
\quad a=1,\dots,A,
\]
where $l_*^{(0)}$ is the reference logit predicted by the GP, and $(\mu_a,\sigma_a^2)$ are annotator-specific bias and variance parameters.

For binary segmentation, we adopt the inverse probit link function
\[
p_{*,a} := p(y_{*,a}=1 \mid l_{*,a}) = \Phi(l_{*,a}),
\]
where $\Phi(x)=\int_{-\infty}^{x}\mathcal{N}(t\mid0,1)\,dt$ denotes the standard normal cumulative distribution function. This yields the closed-form predictive probability \cite{nickisch2008approximations}:
\begin{equation}\label{cls_prob_bias_var}
p_{*,a}
=
\Phi\!\left(
\frac{\tau+\mu_a}{\sqrt{1+\eta^2+\sigma_a^2}}
\right).
\end{equation}

The learnable parameters include the feature extractor parameters $\psi$, kernel hyperparameters $\varphi$, inducing inputs $\mathbf{Z}_u$, variational parameters $(\mathbf{m}_u,\mathbf{S})$, and annotator-specific parameters $\boldsymbol{\mu}=(\mu_1,\dots,\mu_A)^\top$ and $\boldsymbol{\sigma}=(\sigma_1,\dots,\sigma_A)^\top$, which are jointly optimised using stochastic gradient descent. Let $y_{i,a} \in \{0,1\}$ denote the annotation of annotator $a$ at pixel $i$, and let $p_{i,a}$ be the corresponding predicted probability.
The loss function is defined as
\begin{equation}
\mathcal{L}
=
-\sum_{a=1}^{A}\text{BCE}(y_{\cdot,a},p_{\cdot,a})
+
\text{KL}[q(\mathbf{u})||p(\mathbf{u})]
-\sum_{a=1}^{A}\text{Dice}(y_{\cdot,a},p_{\cdot,a})
+
(\sum_{a=1}^{A}\mu_a)^2
\end{equation}
where binary cross-entropy and KL divergence form the ELBO objective \cite{hensman2015scalable}. The soft Dice loss~\cite{eelbode2020optimization} is added empirically to support segmentation training, following common practice in medical image segmentation.

\section{Experiment}
\subsection{Dataset} 
The trans-rectal ultrasound (TRUS) images were acquired using a side firing transducer integrated within a bi-plane trans-perineal ultrasound probe, along with a digital stepper. During acquisition, the stepper was manually positioned at the centre of the anatomical region of interest, and then manually rotated at predefined angular intervals to scan the entire prostate gland. 
From the acquired $249$ TRUS volumes, a total of $6644$ 2D slices were sampled for feasible annotation, with a pixel size of $0.18\times0.16$ mm/pixel and an image size of $361\times403$ pixels. 
 
Each ultrasound image was annotated independently by three researchers (\textit{Annotation 1–3}). In addition, a high-quality reference annotation (\textit{Annotation HQ}) was provided by a clinician. The high-quality annotation was obtained by refining the majority-vote segmentation derived from the three researcher annotations.

\subsection{Implementation Details} 
At the patient level, $149$/$51$/$49$ cases were randomly assigned to the training/validation/test sets, corresponding to $3997$, $1310$, and $1337$ images, respectively.
The feature extractor adopts a standard U-Net architecture with feature dimension $d=64$.
The Gaussian Process module is implemented using GPyTorch~\cite{gardner2018gpytorch} with $M=512$ inducing points and RBF kernel. Variance parameters are constrained to remain positive.

Models are optimised using the AdamW optimiser with a learning rate of $10^{-4}$. Hyperparameters are selected based on validation performance.
During training, for each image, one annotation is randomly sampled to compute the loss, with the corresponding $\mu_a$ and $\sigma_a$ applied (for annotation HQ, $\mu_a=0$ and $\sigma_a=0$). During inference, annotator-specific predictions are obtained using Eq.~\eqref{cls_prob_bias_var} with the learned bias and variance parameters.
All models are trained on a single NVIDIA A100 GPU for up to one week until convergence. 
The GP module requires less than 0.4 seconds per image during training and 0.3 seconds per image during inference.

\subsection{Comparison Methods}
We compare our approach with two baselines. First, we train separate U-Net models for each annotator using their respective annotations as supervision. This setup evaluates the performance achievable when annotator disagreement is not explicitly modelled, but instead implicitly absorbed into independently trained models.

We also compare with Pionono~\cite{schmidt2023probabilistic}, a state-of-the-art probabilistic multi-rater segmentation framework that captures inter- and intra-observer variability through annotator-specific latent distributions learned via variational inference. Pionono produces annotator-conditioned probabilistic segmentations by Monte-Carlo sampling from these latent posteriors.

Evaluation metrics. The segmentation accuracy is evaluated by Dice score (Dice) and 95th Percentile Hausdorff Distance (HD95) \cite{muller2022towards}. The calibration quality is measured by Expected Calibration Error (ECE) \cite{guo2017calibration} and Negative Log-Likelihood (NLL) \cite{gneiting2007strictly}.

\section{Results and Discussion}

Table~\ref{tab1} summarises the segmentation and calibration performance of the proposed SVGP model compared with the individual U-Net baselines and Pionono, evaluated on the test set. For each of Annotation 1–3 and HQ, annotator-conditioned predictions are evaluated against the corresponding annotation. All metrics are averaged over test images.

\begin{table}[!t]
\caption{Segmentation and calibration performance of U-Net, Pionono, and the proposed SVGP model on the test set. Annotator-conditioned predictions are evaluated against the corresponding annotation. Boldface indicates the best performance. An asterisk denotes a statistically significant difference ($p<0.05$) relative to SVGP.}\label{tab1}
\centering
\footnotesize
\begin{tabular*}{\linewidth}{@{\extracolsep{\fill}} c|c|c|c|c|c}
\hline
Annotator &  Model & ECE $\downarrow$ & NLL $\downarrow$  & Dice $\uparrow$    & HD95 (mm) $\downarrow$   \\
\hline

\multirow{3}{*}{HQ}
& U-Net & $0.029\pm0.036^{*}$ &  $0.292\pm0.457^{*}$ &  $0.840\pm0.247$ &   $5.742\pm4.762^{*}$    \\ 
& Pionono &     $0.031\pm0.038^{*}$ &  $0.464\pm0.609^{*}$  & $\mathbf{0.850\pm0.239}$ &  $\mathbf{5.218\pm4.863}$ \\
& Ours &  $\mathbf{0.025\pm0.035}$ &  $\mathbf{0.141\pm0.328}$ & $0.843\pm0.241$ &  $5.298\pm4.621$  \\
\hline
\multirow{3}{*}{1}
& U-Net & $0.036\pm0.046^{*}$ &  $0.397\pm0.619^{*}$ &$0.797\pm0.286^{*}$ &   $6.311\pm4.805^{*}$    \\
& Pionono &  $0.036\pm0.046^{*}$ &  $0.552\pm0.730^{*}$ & $0.815\pm0.272$ &   $\mathbf{5.819\pm5.244}$  \\
& Ours &  $\mathbf{0.030\pm0.044}$ &  $\mathbf{0.173\pm0.406}$ & $\mathbf{0.817\pm0.264}$ &   $5.850\pm4.931$  \\

\hline
\multirow{3}{*}{2}
& U-Net &  $0.053\pm0.057^{*}$ &  $0.578\pm0.683^{*}$ & $0.678\pm0.361^{*}$ &  $8.698\pm5.394^{*}$  \\
& Pionono &  $0.049\pm0.068^{*}$ &  $0.753\pm1.068^{*}$ & $0.733\pm0.359$ &  $\mathbf{5.730\pm4.714}^{*}$  \\
& Ours &  $\mathbf{0.040\pm0.051}$ &  $\mathbf{0.146\pm0.187}$ & $\mathbf{0.736\pm0.346}$ &    $6.417\pm3.747$  \\
\hline
\multirow{3}{*}{3}
& U-Net &  $0.046\pm0.049^{*}$ &  $0.421\pm0.570^{*}$ & $0.762\pm0.287^{*}$ &   $7.511\pm4.923^{*}$  \\
& Pionono &  $0.043\pm0.048^{*}$ &  $0.651\pm0.752^{*}$ & $\mathbf{0.811\pm0.263}$ &  $\mathbf{6.172\pm4.863}$  \\
& Ours &  $\mathbf{0.036\pm0.043}$ &  $\mathbf{0.203\pm0.363}$ & $0.804\pm0.262$ &   $6.444\pm4.818$  \\

\hline
\end{tabular*}
\end{table}

Across all annotators, the individual U-Net models obtain the lowest Dice and the highest HD95, indicating that independently training separate models for each annotator provides a limited characterisation of annotation variability. Pionono achieves the highest Dice in some cases and the lowest HD95 in all cases, indicating strong segmentation accuracy. However, its NLL is consistently higher than U-Net, revealing that gains in segmentation accuracy do not correspond to improved probability calibration. 

In contrast, the proposed SVGP model achieves the lowest ECE and NLL across all annotators. The reduction in NLL is particularly pronounced for Annotator 2, where it decreases from 0.578 (U-Net) and 0.753 (Pionono) to 0.146. Notably, these large improvements in ECE and NLL are achieved while maintaining Dice and HD95 values comparable to Pionono, demonstrating that the proposed model significantly improves calibration without sacrificing segmentation performance.

The learned annotator-specific parameters are
\[
(\mu_1,\sigma_1^2)=(0.265,0.208), (\mu_2,\sigma_2^2)=(-1.042,0.946), (\mu_3,\sigma_3^2)=(0.772,0.738)
\]
with the reference logit distribution corresponding to $(0,0)$.
The magnitude of $\sigma_a$ matches the performance of the corresponding deterministic individual U-Net: larger $\sigma_a$ coincides with worse metrics. Each deterministic individual U-Net is trained to match a single annotator. When those labels exhibit higher intra-rater variability, the supervision signal becomes less consistent, making the mapping harder to learn. The observed correspondence therefore suggests that the learned variance parameter captures this intra-rater variability quantitatively.

\begin{figure}[!t]
\includegraphics[width=\textwidth]{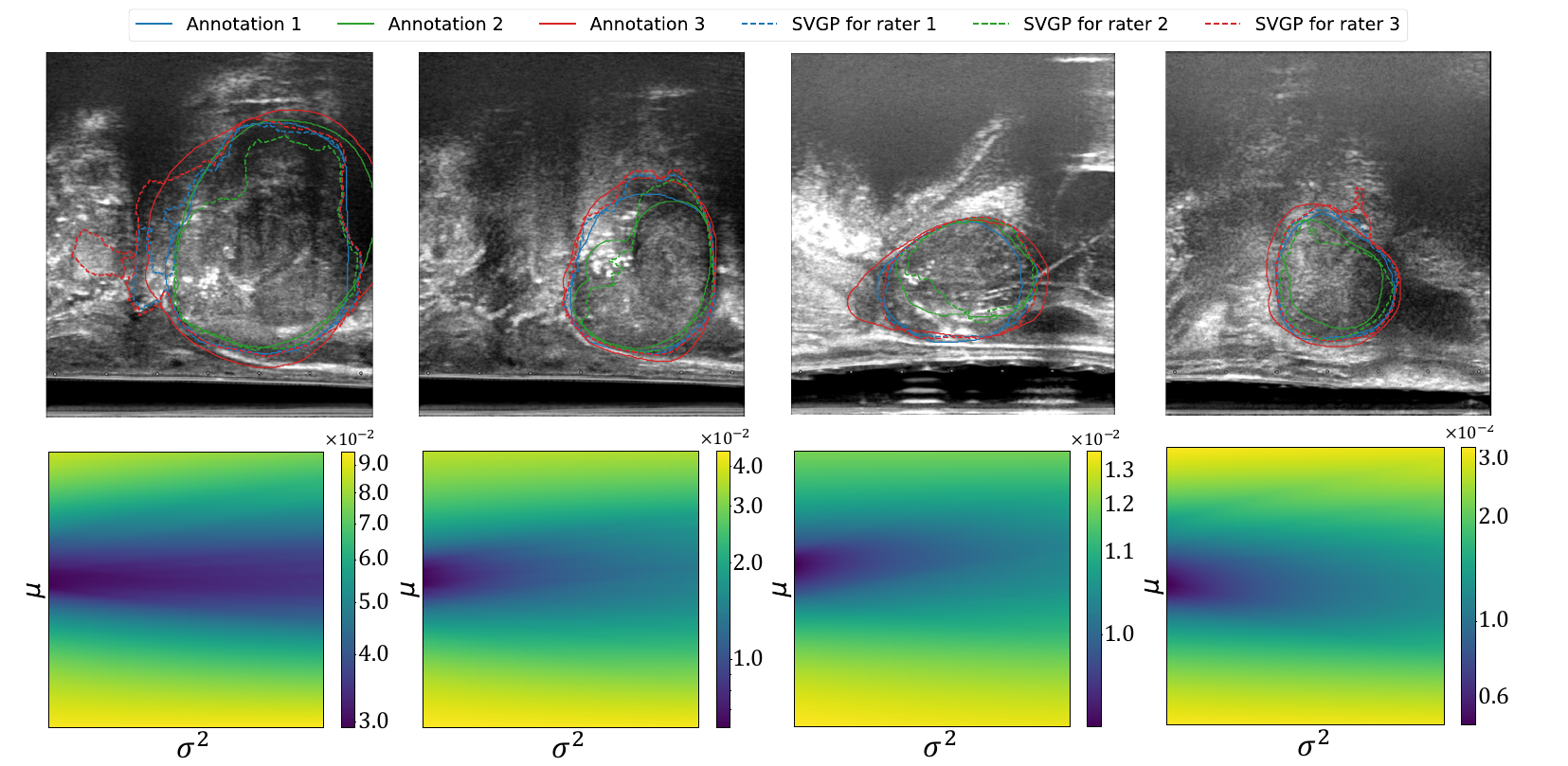}
\caption{Top row: SVGP-based predictions and the corresponding annotations for each rater. Bottom row: predictive ECE evaluated over a grid of bias and variance parameters ($\mu \in [-3.0, 3.0]$, $\sigma^2 \in [0.1, 3.0]$)} \label{pred_ece}
\end{figure}

\textbf{Visualization of the relationship between predictive performance and bias and variance.}
The top row of Fig.~\ref{pred_ece} presents the annotations and the corresponding SVGP-based predictions for each rater. The predicted segmentations closely resemble the respective annotator labels, indicating that the proposed formulation effectively captures inter-rater variability through the learned perturbation parameters.

The bottom row visualizes the ECE for each image, computed by simulating predictions using Eq.~\eqref{cls_prob_bias_var} over a grid of bias and variance values ($\mu \in [-3.0, 3.0]$, $\sigma^2 \in [0.1, 3.0]$) and comparing them with Annotation HQ. By systematically perturbing the reference logit distribution, this experiment examines how annotation bias and variability influence predictive calibration relative to a common reference.

The observed trends indicate that ECE is considerably more sensitive to changes in bias $\mu$ than to changes in variance $\sigma^2$. Systematic logit shifts lead to pronounced calibration degradation, whereas increased variance produces comparatively moderate effects. This suggests that inter-rater disagreement, reflected as systematic bias, has a stronger impact on predictive reliability than within-rater variability.

\section{Conclusion}

We proposed a logit-space SVGP formulation that explicitly parameterises annotator dependent bias and variance, providing an interpretable framework for modelling annotation-induced variability. By separating image-dependent predictions from annotation perturbations, the approach enables direct analysis of how inter-rater bias and intra-rater variability influence predictive uncertainty and downstream performance.

The formulation allows controlled examination of the sensitivity of segmentation accuracy and calibration to different noise components, offering practical insight for dataset curation. Under fixed labelling budgets, it can help assess whether reducing systematic inter-rater bias or addressing intra-rater variability is more likely to improve predictive reliability.

While the current evaluation demonstrates the feasibility of the framework in a clinically relevant multi-rater setting, future work will further validate its generalisability on additional datasets and extend the annotator model beyond global rater-level bias and variability to capture image- and structure-dependent disagreement with spatial correlations, as well as multi-class segmentation.

\begin{credits}
\subsubsection{\ackname} 

This work was supported by the EPSRC [EP/T029404/1], a Royal Academy of Engineering/Medtronic Research Chair [RCSRF1819\textbackslash7\textbackslash734] (TV), Wellcome/EPSRC Centre for Interventional and Surgical Sciences [203145Z/16/Z], and the International Alliance for Cancer Early Detection, an alliance between Cancer Research UK [C28070/A30912; C73666/A31378; EDDAPA-2024/100014], Canary Center at Stanford University, the University of Cambridge, OHSU Knight Cancer Institute, University College London and the University of Manchester. TV is co-founder and shareholder of Hypervision Surgical. Qi Li was supported by the University College London Overseas and Graduate Research Scholarships.

\subsubsection{\discintname}
The authors have no competing interests to declare that are
relevant to the content of this article. 
\end{credits}


\bibliographystyle{splncs04}
\bibliography{ref}

\end{document}